\documentclass[a4paper,twoside]{article}

\usepackage{adjustbox}
\usepackage{caption}
\usepackage{hyperref}
\usepackage{comment}
\usepackage{titlesec}
\usepackage{enumitem}

\usepackage{epsfig}
\usepackage{subcaption}
\usepackage{calc}
\usepackage{amssymb}
\usepackage{amstext}
\usepackage{amsmath}
\usepackage{amsthm}
\usepackage{multicol}
\usepackage{pslatex}
\usepackage{apalike}

\usepackage{SCITEPRESS}

\begin{document}

\title{A novel method for object detection using deep learning and CAD models}

\author{\authorname{Igor Garcia Ballhausen Sampaio\sup{1}, Luigy Machaca\sup{1}, José Viterbo\sup{1}\orcidAuthor{0000-0002-0339-6624} and Joris Guérin\sup{2}\orcidAuthor{0000-0002-8048-8960}} \affiliation{\sup{1}Computing Institute, Universidade Federal Fluminense, Brazil} \affiliation{\sup{2}LAAS-CNRS, ONERA, Université de Toulouse, France} \email{\{igorgarcia, luigyarcana, jviterbo\}@id.uff.br, jorisguerin.research@gmail.com}}


\abstract{Object Detection (OD) is an important computer vision problem for industry, which can be used for quality control in the production lines, among other applications. Recently, Deep Learning (DL) methods have enabled practitioners to train OD models performing well on complex real world images. However, the adoption of these models in industry is still limited by the difficulty and the significant cost of collecting high quality training datasets. On the other hand, when applying OD to the context of production lines, CAD models of the objects to be detected are often available. In this paper, we introduce a fully automated method that uses a CAD model of an object and returns a fully trained OD model for detecting this object. To do this, we created a Blender script that generates realistic labeled datasets of images containing the object, which are then used for training the OD model. The method is validated experimentally on two practical examples, showing that this approach can generate OD models performing well on real images, while being trained only on synthetic images. The proposed method has potential to facilitate the adoption of object detection models in industry as it is easy to adapt for new objects and highly flexible. Hence, it can result in significant costs reduction, gains in productivity and improved products quality.}

\maketitle
\setcounter{footnote}{0}
\normalsize

\section{\uppercase{Introduction}}
\label{sec:introduction}

Recently, Deep Learning (DL) has produced excellent results for Object Detection (OD)~\cite{liu2020deep}. On the one hand, a typical limitation with DL is the requirement of large labeled datasets for training. Indeed, although there are various large databases available online for OD, for specific industrial applications it is always necessary to create custom datasets containing the objects of interest. While the scenarios present in public datasets are useful from both a research and application standpoint, it was found that industrial applications, such as bin picking or defect inspection, have quite different characteristics that are not modeled by the existing datasets~\cite{drost2017introducing}. As a result, methods that perform well on existing datasets sometimes show different results when applied to industrial scenarios without retraining. The process of generating a specific dataset for retraining is tedious, and can be error-prone when conducted by non-professional technicians. Moreover, generating a new dataset and labeling it manually can be very time consuming and expansive~\cite{jabbar2017training}.

On the other hand, OD for industrial production lines presents the specificity that the manufacturers often have access to the CAD models of the objects to detect. Thanks to advances in computer graphics techniques, such as ray tracing~\cite{shirley2003realistic}, the generation of photo-realistic images is now possible. In such artificially generated images, the computer can be employed to obtain bounding box labeling for free. The use of synthetic images rendered from CAD models to train OD models has already been proposed in \cite{peng2015learning}, \cite{rajpura2017object} and \cite{hinterstoisser2018pre}. However, their approaches are not automated as they require manual scene creation by Blender artists. In addition, the objects used in these works are usually generic, such as buses, airplanes, cars or animals.

The main contribution of this paper is to present a new method for training OD models in synthetic images generated from CAD models that is fully automatic and thus well suited for industrial use. The proposed method consists of the automatic generation of realistic labeled images containing the objects to be detected, followed by the fine-tuning of a pretrained OD model on the artificial dataset. An extensive study is conducted to properly select the user-defined parameters so that it maximizes the performance on real world images. Our method is evaluated using the CAD models of two industrial objects for training, as well as real labeled images containing the objects for evaluation. The results obtained are very promising as we manage to get F1-scores above 90\% on real images while training only on synthetic images.

This paper is organized as follows. Section~\ref{sec:related_work} presents the related work in the field of object detection and deep learning for industry. Section~\ref{sec:methodology} provides detailed explanations about the proposed method created. Section~\ref{sec:results} describes our experiments, presents the results obtained and discusses them. Finally, conclusions and directions for future work are presented in Section~\ref{sec:conclusion}.

\section{\uppercase{RELATED WORK}}
\label{sec:related_work}

This section presents related work about OD, industrial applications of DL-based computer vision, as well as computer vision methods using CAD models.

\subsection{Object Detection}

OD is a challenging computer vision problem that consists in locating instances of objects from predefined categories in natural images~\cite{od_survey_old}. It has many applications in various domains such as autonomous driving, security and medical diagnosis~\cite{xiao2020review}. Deep learning techniques have emerged as a powerful strategy for learning characteristic representations directly from data and have led to significant advances in the field of generic object detection~\cite{liu2020deep}. In the last decade, many competitions for object detection have been held to provide large annotated datasets to the community, and to unify the benchmarks and metrics for fair comparison between proposed methods~\cite{everingham2010pascal}, \cite{lin2014microsoft}, \cite{zhou2017places}, \cite{kuznetsova2018open}.

Some examples of OD methods proposed within the last few years include~\cite{he2015spatial}, where the author proposes a new network structure, called SPP-net, which can generate a fixed-length representation, regardless of the size/scale of the image. Other works such as~\cite{jana2018yolo} aim to improve processing speed and at the same time efficiently identify objects in the image. Finally, deeper CNNs have led to record-breaking improvements in the detection of more general object categories, a shift which came about when DCNNs began to be successfully applied to image classification~\cite{liu2020deep}.

\subsection{OD for Industrial Applications}

Although general purpose OD methods have greatly improved thanks to the availability of large public datasets, the detection of instances in the industrial context must be approached differently, since annotated images are generally not available or rare. Indeed, to train a deep learning model, hundreds of annotated images for each object category are needed. Specific datasets need to be collected and annotated for different target applications. This process is time-consuming and laborious, and increases the burden on operators, which goes against the goal of industrial automation~\cite{cohen2020cad}, \cite{ge2020towards}.

A public datasets adapted to the industrial context was developed in~\cite{drost2017introducing}. Unlike other 3D object detection datasets, this work models industrial waste collection and object inspection tasks that often face different challenges. In addition, the evaluation criteria are focused on practical aspects, such as execution times, memory consumption, useful measures of correction and precision. Other examples of datasets adapted to the industrial context include~\cite{guerin2018semantically} and \cite{guerin2018automatic}.

Finally, in~\cite{yang2019real}, a method to detect defects of tiny parts in real time was developed, based on object detection and deep learning. To improve their results, the authors consider the specificities of the industrial application in their method such as the properties of the parts, the environmental parameters and the speed of movement of the conveyor. This is a good example to adapt OD training methods to the specific constraints of the industrial context.

\subsection{CAD Models and OD}

The first commercial CAD programs came up in the 1970s, providing functions for 2D-drawing and data archival, and evolved into the main engineering design tool \cite{lindsay2018identifying}, \cite{hirz2017future}. These models can provide a scalable solution for intelligent and automatic object recognition, tracking and augmentation based on generic object models \cite{ben2010computer}. For example, CAD models have been used to support multi-view detection~\cite{zhang2013object}. In \cite{peng2015learning}, 3D models were used as the primary source of information to build object models. In other works, 3D CAD models were used as the only source of labeled data~\cite{lin2014microsoft}, \cite{everingham2010pascal}, but they are limited to generic categories, such as cars and motorcycles .

\begin{figure*}[!ht]

\centering
\vspace{10pt}

\subfloat[][Training]{ \includegraphics[width=0.98\textwidth]{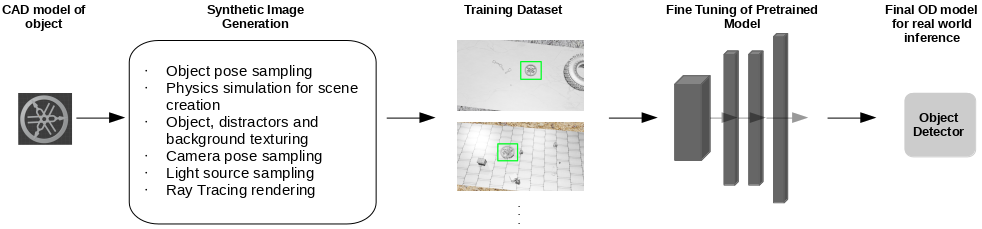}\label{fig:training}}

\subfloat[][Inference]{\includegraphics[width=0.6\textwidth]{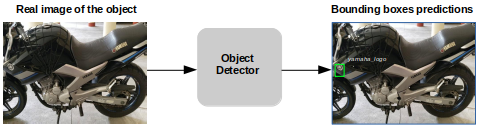}\label{fig:inference}}

\caption{Overview of the proposed method for training an object detection network using a CAD model.}
\label{fig:methodology}
\end{figure*}

\section{\uppercase{PROPOSED METHOD}}
\label{sec:methodology}

An overview of this proposed method can be seen in Figure~\ref{fig:methodology}. First, a custom Blender code is used to generate labeled training images containing the rendered CAD model in context. Then, a pretrained object detection model is fine-tuned on the generated dataset. Finally, the model can be used for inference on real images (Figure~\ref{fig:inference}).

\subsection{Image Generation} \label{sec:method_ImGen}

For the automatic generation of the training images, the software Blender~\cite{blender2018blender} is used. Blender is a powerful software for 3D design, which includes features such as modeling, rigging, simulation and rendering. Blender has a good Python API, is open-source and has good GPU support.

\begin{figure}[tb]
     \centering
     \begin{subfigure}[b]{0.2\textwidth}
         \centering
         \includegraphics[width=1\linewidth]{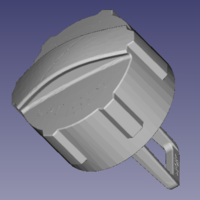}
         \caption{Adblue}
         \label{fig:CAD_adblue}
     \end{subfigure}
     \hfill
     \begin{subfigure}[b]{0.2\textwidth}
         \centering
         \includegraphics[width=1\linewidth]{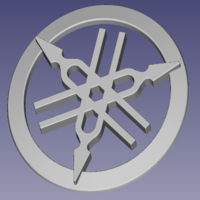}
         \caption{Yamaha logo}
         \label{fig:CAD_yamaha}
     \end{subfigure}
        \caption{CAD models used in our experiments.}
        \label{fig:cad_models}
\end{figure}

In order to generate a synthetic training image sample, our code requires several elements. First, a CAD model of the object of interest as well as several other industrial CAD models need to be available. In the experiments of this paper, we use the two objects shown in Figure~\ref{fig:cad_models}, for which we also have real world test images. The other objects serve as distractors to help the model focusing on the right object. The CAD models for the distractors are gathered from the Grabcad website\footnote{\url{https://grabcad.com/}}. Different textures for the different distractors as well as for the background are gathered from the Poliigon website\footnote{\url{https://www.poliigon.com/}}. Finally, the color and texture of the object of interest are reproduced manually.

\begin{figure*}[ht]
     \centering
     \begin{subfigure}[b]{0.48\textwidth}
         \centering
         \includegraphics[width=\textwidth]{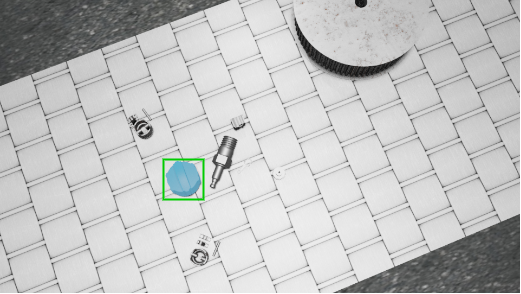}
         \caption{Adblue}
         \label{fig:example-blender-adblue}
     \end{subfigure}
~
     \begin{subfigure}[b]{0.48\textwidth}
         \centering
         \includegraphics[width=\textwidth]{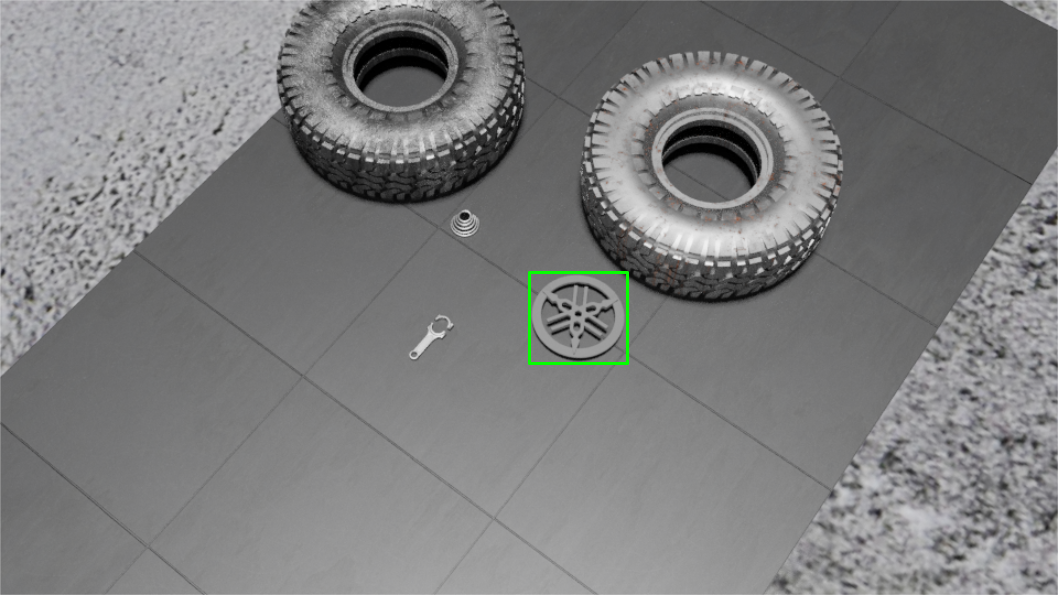}
         \caption{Yamaha logo}
         \label{fig:example-blender-yamaha}
     \end{subfigure}
        \caption{Example of images generated using our custom Blender script.}
        \label{fig:example-blender}
\end{figure*}

Once we have access to all the elements above, the generation code goes as follows. A floor and a table are created and some distractors are sampled. Using physics simulation, the distractors are dropped from a random height on the table. The position of the object of interest is also randomly sampled. Once the 3D scene is created, textures and colors are sampled for the backgrounds and the distractors and the entire scene is textured. Light sources and cameras are also sampled and placed randomly. Constraints on the camera pose are applied, in order to ensure that the object appears in the camera view. Once the scene has been created, the rendering occurs and generates an image. By removing the light sources and making the object of interest a light source itself, we can generate another image which can be used for bounding box labeling. This procedure is necessary because even if we know the location of the object, it can be partly hidden by distractors and thus distort the labeling. Example images generated using our Blender code can be seen in Figure~\ref{fig:example-blender}.

\subsection{Model Training}

In this work, we did not train a new CNN architecture from the scratch. Instead, we used one of the pre-trained models provided by TensorFlow Object Detection API~\cite{TF_OD_API}. This approach is called transfer learning and consists in starting training from a model that already knows basic feature extraction skills and is less likely to overfit the synthetic datasets. Indeed, the diversity that we can create with Blender is limited as we cannot get an infinite amount of textures and distractors, and the diversity already encountered by the network during pre-training can help reduce overfitting. In addition, using a network pre-trained on real images can prevent the network from learning detection features that depend too much on the generation procedure.

There exists several models in the TensorFlow OD model zoo. More information on the performance of the detection, as well as the reference execution times, for each of the available pre-trained models, can be found on the Github page of the API\footnote{\url{https://github.com/tensorflow/models}}. In practice, the model used in this paper is the \textit{faster\_rcnn\_inception\_v2\_coco} model, which provides a good trade-off between performance and speed.

Faster R-CNN, the model used in this work, takes as input an entire image and a set of object proposals. The network first processes the whole image with several convolutional and max pooling layers to produce a convolutional feature map. Then, for each object proposal, a region of interest (RoI) pooling layer extracts a fixed-length feature vector from the feature map. 
Each feature vector is fed into a sequence of fully connected layers that finally branch into two sibling output layers: one that produces softmax probability estimates over \textit{K} object classes plus a catch-all “background” class and another layer that outputs four real-valued numbers for each of the \textit{K} object classes. Each set of 4 values encodes refined bounding-box positions for one of the \textit{K} classes. For a more detailed view about Faster-RCNN, we refer the reader to the original paper~\cite{ren2015faster}, or to the following tutorial~\cite{tuto_fasterrcnn}.

To train the final OD model, the TensorFlow OD API requires a specific file structure of the training images and labels. This step is carried out automatically by our script.

\subsection{Parameter Selection Procedure}

The Blender script used for image generation has many hyperparameters that must be chosen before using it, such as the number of distractors, the number of scenes generated or the resolution of synthetic images. Hence, we conduct a set of experiments to properly select these parameters in order to optimize the OD results for inference on real images. In this section, we explain the parameter selection procedure. In other words, we present the dataset on which the different sets of parameters were evaluated, as well as the metrics used to assess the quality of the results obtained with a given set of parameters. The results obtained for this parameter selection procedure are presented in Section~\ref{sec:results}.

\subsubsection{Test dataset}
The objective of this work is to validate that an object detector trained on synthetic images can generalize to real world industrial cases. Hence, we use a test dataset composed of 380 real images containing the objects corresponding to the CAD models used for training. The bounding box annotation files for the test images are generated manually using a software called LabelImg\footnote{\url{https://github.com/tzutalin/labelImg}}. This application allows us to draw and save the annotations of each image as xml files in the PASCAL VOC format~\cite{everingham2010pascal}.

\subsubsection{OD metrics}
In order to evaluate the quality of the trained model on real images, and thus to be able to select the best hyperparameters for image generation and training, we used standard OD metrics that are presented here.

\begin{figure}[!t]
\centering
\includegraphics[width=0.45\textwidth]{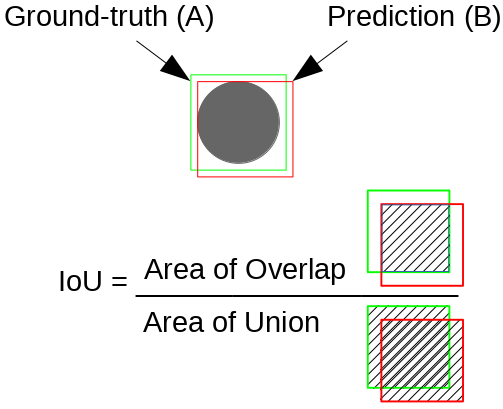}
\caption{Intersection over Union (IoU) computation}
\label{object_bb}
\end{figure}

\paragraph{Intersection over Union (IoU)} is an evaluation metric used to measure how much a predicted bounding box matches with a ground truth bounding box. For a pair of bounding boxes, IoU is defined as the area of the intersection divided by the area of the union (Figure~\ref{object_bb}). If A corresponds to the ground-truth box and B to the predicted box, then, IoU is computed as :
\begin{equation}
IoU = \frac{|A\cap B|}{|A\cup  B|},
\end{equation}
where $|.|$ denotes the area of a given shape. The numerator is called the overlap area and the denominator is called the combined area. IoU ranges between 0 and 1, where 1 means that the bounding boxes are the same and 0 that there is no overlap.

\paragraph{Precision, Recall, F1-Measure}

We call confidence score, the probability that an anchor box contains an object from a certain class. It is usually predicted by the classifier part of the object detector. The confidence score and IoU are used as the criteria to determine whether a detection is a true positive or a false positive. Given a minimal threshold on the confidence score for bounding box acceptance, and another threshold on IoU to identify matching boxes, a detection is considered a true positive (TP) if there exists a ground truth such that: confidence score $>$ threshold; the predicted class matches the class of the ground truth; and $\text{IoU}> \text{threshold}_{\text{IoU}}$. The violation of any of the last two conditions generates a false positive (FP). In case multiple predictions correspond to the same ground-truth, only the one with the highest confidence score counts as a true positive, while the others are considered false positives. When a ground truth bounding box is left without any matching predicted detection, it counts as a false negative (FN).

If we note TP, FP and TN respectively the number of True Positives, False Positives and False Negatives in a dataset, we can define the following metrics:
\begin{equation}
Precision = \frac{TP}{TP+FP},
\end{equation}
\begin{equation}
Recall = \frac{TP}{TP+FN},
\end{equation}
\begin{equation}
F_1 = \frac{Precision \cdot Recall }{Precision + Recall}.
\end{equation}
A high precision means that most of the predicted boxes had a corresponding ground truth, i.e., the object detector is not producing bad predictions. A high recall means that most of the ground truth boxes had a corresponding prediction, i.e., the object detector finds most objects in the images. The $F_1$-Score is the harmonic mean of the precision and recall, it is needed when a balance between precision and recall is sought.

In the case of object detection on production lines, a low precision means that sometimes a part might be absent and the model would not see it, whereas a low recall means that sometimes the part is present and the model raises an alert anyways. For this reason, both a good recall and a precision are required and the choice of using the $F_1$-Score metric seems appropriate.

\paragraph{Average Precision}
After an OD model has been trained, the computation of Precision, Recall and F1-score depends on the value of the two thresholds defined above (for the confidence score and IoU). In order to properly choose the values of these thresholds, it is interesting to analyze the Precision x Recall curves. For each class, and for a given value of the IoU threshold, the confidence threshold is set as a variable and sampled between 0 and 1 to plot a parametric curve with precision and recall as the x and y-axis.

A class-specific object detector is considered good if the precision remains high as the recall increases, meaning that if you vary the confidence limit, the precision and recall will still be high. Hence, to compare between curves we generally rely on a numerical metric called Average Precision (AP). Since 2010, the standard computation method for AP consists in calculating the area under the curve (AUC) of the Precision x Recall curve~\cite{everingham2010pascal}.

\section{\uppercase{EXPERIMENTS AND RESULTS}}
\label{sec:results}

The results obtained for the parameter selection procedure as well as our final evaluations are presented here. These experiments were conducted on a Nvidia Quadro P5000 GPU and a 2.90GHz Intel Xeon E3-154M v5 processor (16 GB of RAM).

\subsection{Hyperparameter tuning}

The parameter selection procedure is conducted exclusively on the Yamaha logo object, the best set of parameters is then tested on the Adblue object to ensure that it also performs well. The influence of four tunable parameters on the final results is studied here. for each parameter, three values were selected for the tests. These parameters and their studied values are:

\begin{itemize}
    \item \textbf{Resolution: } 640x480, 960x540, 1080x720
    \item \textbf{Camera poses: } 2, 5, 20
    \item \textbf{Number of scenes: } 20, 50, 200
    \item \textbf{Number of distractors: } 0, 5, 20
\end{itemize}
From simple preliminary experiments that are not presented here, we concluded that the number of textures used for the floor, the distractors and the support should be set to the maximum number of textures available (in our case 7 for the floor and 6 for the two others). The parameter values used in this work were chosen empirically, that is, after several test scenarios, these values were the ones that generated the best performance regarding the metrics.

In total, from the values selected for the four parameters, we sampled more than 30 combinations and compared the OD results on the testing set of real images. For each combination tested, we trained the Faster-RCNN CNN on the synthetic images that were generated. We note that, for each hyperparameters combination, the experiments were repeated 10 times in order to attenuate the influence of the random components in the generation and training process. For reasons of space in this article, it was not possible to present all results. However, in order to demonstrate the importance of this parameter selection step, Table~\ref{tab:best_worst} shows the best and the worst configuration that were tested.

\begin{table}[!t]
\centering
\caption{Best and worst hyperparameters configurations obtained and their corresponding results.}
\label{tab:best_worst}
\begin{adjustbox}{width=0.45\textwidth}
\begin{tabular}{|c|c|c|}
\hline
\textbf{Parameters} & \textbf{Best Case} & \textbf{Worst Case}  \\
\hline
Resolution & 960x540 & 960x540 \\
\hline
cam\_poses & 5 & 5  \\
\hline
n\_scenes              & 20        & 20         \\
\hline
n\_images              & 100       & 100         \\
\hline
n\_distractors         & 20        & 0           \\
\hline
Generation Time        & 1257.30   & 749.92      \\
\hline
n\_samples             & 10        & 10          \\
\hline \hline
Precision \%: Avg \textit{(Std.Dev)} & 78.06 \textit{(15.41)} & 57.34 \textit{(16.27)} \\
\hline
Recall \%: Avg \textit{(Std.Dev)} & 96.223 \textit{(4.07)} & 90.71 \textit{(6.45)}       \\
\hline
F1-Score \%: Avg \textit{(Std.Dev)} & 85.19 \textit{(10.57)} & 66.80 \textit{(11.49)}     \\ \hline
\end{tabular}
\end{adjustbox}
\end{table}

In Table~\ref{tab:best_worst}, we can see that the distractors are an essential element in our proposed pipeline for image generation. Indeed, when removing them, we can see a drop of around 23\% in F1-score, on average across the 10 experiment samples. We also tried combinations with few distractors, but the F1-score results dropped significantly. This makes sense as the real images evaluated had several distractors as well.

Another important point is that the resolution of the images generated should be greater than the inference images. In all of our tests, this scenario always produced the best results. It also makes sense as it is easier to learn from a more detailed/complex model and then evaluate in a less detailed/complex scenario.

Finally, we also tried to increase the number of generated training images to see if this would lead to an increase in performance. Surprisingly, we acknowledged that the performance dropped for the case with 20 distractors, 20 camera poses and 50 scenes (1000 images). This might mean that when presented too many synthetic images, the model starts overfitting to the biases involved by our generation process, and it also indicates that we do not need a large number of images to train our model. In addition to this performance drop, generating ten times more images also makes the proposed pipeline almost 25 times slower (31037.16 seconds).

\subsection{Results}

In this section, we evaluate the results of the best combination of parameters (Best Case from Table~\ref{tab:best_worst}) in more details. These results are presented in Table~\ref{results_1}, they correspond to using a confidence threshold of 0.9 and an IoU threshold of 0.5. From Table~\ref{results_1}, we can see that the best parameters identified using only the Yamaha logo produce similar results when applied to another object (Adblue). This suggests that the proposed parameters for our method seem to be well suited for different objects and thus could generalize well to various industrial use cases without additional parameter tuning.

\begin{table}[t]
\centering
\caption{Results obtained with the best set of hyperparameters}
\label{results_1}
\begin{adjustbox}{width=0.45\textwidth}
\begin{tabular}{|c|c|c|c|c|}
\hline
\textbf{Object} & \textbf{Precision \%} & \textbf{Recall \%} & \textbf{F1-Score \%}  \\
\hline
Adblue & 85.11 & 80.00 & 81.93 \\
\hline
Yamaha logo  & 78.06 & 96.22 & 85.19 \\
\hline
\end{tabular}
\end{adjustbox}
\end{table}

\subsection{Discussion} \label{sec:discussion}

It is difficult to compare our results with other works in the literature. Indeed, as far as we know, the approach presented in this work is the first proposal to build a fully automated pipeline that takes as input the CAD model of an object and outputs a trained object detection model for this object without any real image. For fair comparison we would need to compare our work with other end-to-end systematic approaches to build OD models from CAD models, which is impossible as it does not exists. Else, we hope that the results presented in this work can serve as a good baseline for comparison of future works in this research direction.

However, we give a rough comparison with other relevant works to give an idea of how well our approach is performing. In~\cite{mazzetto2019automatic} the detection of objects in an automobile production line was implemented, using only real images of the objects. In this work, the estimated detection accuracy was around 90~\%, which is only about 5 to 10~\% better than the results obtained in our work using only synthetic images. In~\cite{jabbar2017training}, the authors also train an OD model using synthetic images generated with Blender and evaluate the results in real images. However, this approach is not entirely automated since the scenes are created manually by Blender artists to ensure photo-realism. The object used in this work for evaluation is a glass of wine and the maximum AP obtained is 71.14~\%. We can see that our systematic approach seems to work better than this approach, however, we cannot reproduce the method on our objects, as we cannot create the scenes manually in the same way that they would. The potential better performance of our approach can be explained by the fact that the loss of photo-realism can be compensated by the higher number of images in our synthetic datasets. Indeed, with our fully automated approach, it is faster and requires no effort to generate more data, unlike in~\cite{jabbar2017training}.

\section{\uppercase{Conclusions}}
\label{sec:conclusion}

This work presents a systematic approach to train object detection models to address industrial scenarios, using only a CAD model of the object of interest as input. The method first generates realistic synthetic images using a custom Blender script, and then trains a faster-RCNN OD model using the TensorFlow OD API. To understand and optimize the different parameters in the proposed pipeline, a systematic parameter selection study is conducted using a Yamaha logo CAD model for training and real images containing the same object in context for evaluation. The selected hyperparameters are then tested on an other object, showing that they can generalize to different scenarios.

Over the last decade, successful deep learning methods have been developed to tackle the challenging problem of generic object detection. However, when it comes to the problem of OD in an industrial environment, the availability of good quality data becomes a bottleneck. To address this issue we proposed to use synthetic images for training, which is challenging as it might not reflect the high variability found in in real industrial environment (objects, pieces and scenery, etc.). In addition, there is also a difficulty in finding CAD models of specific industrial objects so that they can be trained and other approaches can be tested and compared. Thus, as a consequence of this work, a set of data was produced and made publicly available for future research \footnote{\url{https://github.com/igorgbs/systematic\_approach\_cad\_models}}.

Therefore, the main conclusion from this work is that it is possible to train an object detection model on a set of synthetic images generated from CAD models with excellent performance. In addition, it was shown that a large set of images is not needed to obtain a significant result. Our experiments indicate that the proposed rendering process is sufficient to obtain good performances and that the way of building and rendering the scenes is crucial for the final result.

\section*{\uppercase{Acknowledgements}}

{Our work has benefited from the AI Interdisciplinary Institute ANITI. ANITI is funded by the French "Investing for the Future – PIA3" program under the Grant agreement n°ANR-19-PI3A-0004.}

\bibliographystyle{apalike}

\end{document}